\pdfoutput=1

\documentclass[11pt]{article}

\usepackage[]{ACL2023}

\usepackage{times}
\usepackage{latexsym}

\usepackage[T1]{fontenc}

\usepackage[utf8]{inputenc}

\usepackage{microtype}

\usepackage{inconsolata}

\usepackage{graphicx}
\usepackage{makecell}
\usepackage{multirow}
\usepackage{setspace}
\usepackage{booktabs}
\usepackage{threeparttable}
\usepackage{CJKutf8}
\usepackage{enumerate}
\usepackage{amsfonts,amssymb,amsmath} 
\usepackage{color}
\title{Error-Robust Retrieval for Chinese Spelling Check}
\author{Xunjian Yin\,,\;Xinyu Hu\,,\;Jin Jiang\,,\;Xiaojun Wan \\
        Wangxuan Institute of Computer Technology, Peking University \\ Center for Data Science, Peking University \\ The MOE Key Laboratory of Computational Linguistics, Peking University \\
  \texttt{\{xjyin,huxinyu,jiangjin,wanxiaojun\}@pku.edu.cn}}

\begin{document}
\begin{CJK*}{UTF8}{gkai}
\maketitle
\begin{abstract}
Chinese Spelling Check (CSC) aims to detect and correct error tokens in Chinese contexts, which has a wide range of applications. However, it is confronted with the challenges of insufficient annotated data and the issue that previous methods may actually not fully leverage the existing datasets. In this paper, we introduce our plug-and-play retrieval method with error-robust information for Chinese Spelling Check (\textbf{RERIC}), which can be directly applied to existing CSC models. The datastore for retrieval is built completely based on the training data, with elaborate designs according to the characteristics of CSC. Specifically, we employ multimodal representations that fuse phonetic, morphologic, and contextual information in the calculation of query and key during retrieval to enhance robustness against potential errors. Furthermore, in order to better judge the retrieved candidates, the n-gram surrounding the token to be checked is regarded as the value and utilized for specific reranking. The experiment results on the SIGHAN benchmarks demonstrate that our proposed method achieves substantial improvements over existing work.

\end{abstract}

\section{Introduction}
The purpose of Chinese Spelling Check (CSC) is to detect and correct spelling errors in Chinese texts, which often occur between characters with similar pronunciations and morphologies.
The research on CSC is significant since it benefits various NLP tasks, such as speech recognition, optical character recognition, Chinese grammar error correction, and so on.
With the development of deep learning and pretrained language models, great progress has been made in this task \citep{etoori2018automatic,guo2019spelling, zhang2020spelling}. Further, many current works have turned to introducing phonetic and morphologic information \citep{nguyen2020domain, wang2021dynamic, zhang2021correcting}.
They are based on statistics from \citet{liu2011visually} that most Chinese spelling errors are caused by phonetic or morphologic (graphic) similarity.

However, Chinese spelling check is still challenging because it suffers from subtle and diverse errors. Furthermore, we speculate that current methods have not fully utilized the training data, let alone the lack of adequate parallel corpora. As briefly shown in Table \ref{example}, even the superior model REALISE \citep{xu2021read} in existing studies fails to correct the spelling error "设(set)" to "这(this)" though the same problem in similar contexts occurs a few times in the training set. Meanwhile, REALISE cannot deal with the case of "偷(steal)" while the correct context "...人偷了..." often appears. Therefore, to make better use of the existing dataset, we introduce the retrieval-augmented method with our elaborately designed error-robust information (ERI) and reranking mechanism on k-nearest neighbors (KNN).

\begin{table}[]
\setlength{\tabcolsep}{2.5pt}
\resizebox{0.48\textwidth}{!}{%
\begin{tabular}{c|ll}
\toprule
\multirow{7}{*}{\begin{tabular}[c]{@{}c@{}}Error\\ Pair\\ in\\ Training\\ Set\end{tabular}} &
  Input &
  因为\textcolor[rgb]{1,0,0}{设(set)}是校长的工作。 \\
 & Output (correct)      & 因为\textcolor[rgb]{0,0.66,0.47}{这(this)}是校长的工作。                       \\
 & Output (model) & 因为\textcolor[rgb]{0,0,1}{涉(relate)}是校长的工作。                     \\
 & Translation  & Because this is the principal's job.   \\ \cmidrule{2-3} 
 &
  \multirow{3}{*}{\begin{tabular}[c]{@{}c@{}}Samples in\\ training set\end{tabular}} &
  我以为\textcolor[rgb]{1,0,0}{设}(\textcolor[rgb]{0,0.66,0.47}{这})是她主演的电影。 \\
 &              & 我们\textcolor[rgb]{1,0,0}{设}(\textcolor[rgb]{0,0.66,0.47}{这})个周末见面。                           \\
 &              & 大家认为\textcolor[rgb]{1,0,0}{设}(\textcolor[rgb]{0,0.66,0.47}{这})是正常的。 ...                                    \\ \midrule
\multirow{7}{*}{\begin{tabular}[c]{@{}c@{}}Correct\\ Usage\\ in\\ Training\\ Set\end{tabular}} &
  Input &
  旁边的人\textcolor[rgb]{1,0,0}{头(head)}了我的手册。 \\
 & Output (correct)      & 旁边的人\textcolor[rgb]{0,0.66,0.47}{偷(steal)}了我的手册。                     \\
 & Output (model) & 旁边的人\textcolor[rgb]{0,0,1}{投(cast)}了我的手册。                      \\
 & Translation & Someone nearby stole my manual.\\ \cmidrule{2-3} 
 &
  \multirow{3}{*}{\begin{tabular}[c]{@{}c@{}}Samples in\\ training set\end{tabular}} &
  有人\textcolor[rgb]{0,0.66,0.47}{偷}了我的钱包。 \\
 &              & 有个女生\textcolor[rgb]{0,0.66,0.47}{偷}了我的东西。                            \\
 &              & 店里的珠宝都被人\textcolor[rgb]{0,0.66,0.47}{偷}了。 ...                                    \\ \bottomrule
\end{tabular}
}
\caption{Examples of Chinese spelling errors, including inputs, correct outputs, outputs from model REALISE, and related samples in the training set.} 
\label{example}
\end{table}

Retrieval-augmented text generation, a new paradigm known as "open-book exam", can further improve the performance of target tasks by integrating deep learning models with traditional retrieval technologies \citep{guu2020retrieval, weston2018retrieve, gu2018search}. Among them, algorithms based on KNN retrieval always predict tokens with a nearest neighbor classifier over a large datastore of cached examples, using representations from a neural model for similarity search \citep{khandelwal2019generalization, khandelwal2020nearest, kassner2020bert}. And they have been proven effective for many NLP tasks, such as machine translation, language modeling, dialogue generation, and so on. However, by contrast, CSC has some significant differences and difficulties, on the basis of which we propose our corresponding RERIC method.

Above all, we follow the basic pattern of the CSC task and retrieval-augmented text generation to enhance the prediction of each token in inputs. However, both correct and incorrect tokens exist in the input text, which makes it confusing and unreasonable to arbitrarily store the traditional semantic representations of each token for retrieval. As mentioned before, incorrect tokens, namely spelling errors, are often caused by phonetic and morphologic similarity. So we incorporate the phonetic and morphologic information of each token itself into the calculation of the query and key.

More importantly, the contextual information around the target token is elaborately encoded instead of its own semantic representation. We suppose that such phonetic and morphologic information fused with contextual encoding is more robust and insensitive, no matter whether the target token is correct or not. Furthermore, there are many overlaps between each pair of input and output texts in CSC since only a few tokens are incorrect. So we propose to store the n-gram surrounding the target token as the value to construct the datastore for further reranking instead of conventionally just storing the token itself. The retrieved candidate is more likely to be the right one if it contains the n-gram that overlaps more with the corresponding positions in the input.

To sum up, we retrieve the k nearest neighbors of the target token with its fused and error-robust information. Then we rerank them based on the extent of n-gram overlap and finally obtain the word distribution through the specific calculation. We introduce our RERIC method above into the pretrained CSC model by linearly interpolating the original word distribution with that from retrieval. The experimental results of RERIC on commonly used SIGHAN benchmarks surpass those of the previous methods. And following ablation studies show that conventional retrieval-augmented methods on CSC will result in failure, verifying the effectiveness and necessity of our proposed error-robust information and reranking mechanism. Furthermore, we can easily expand relevant data by directly adding non-parallel texts to the datastore, unlike other works that need to struggle with constructing pseudo-data. Our contributions are as follows:

\begin{enumerate}[1)]
\item To our best knowledge, our work is the first to employ the retrieval-augmented method on the Chinese spelling check task, which can be used in a plug-and-play manner without training and allows more flexible expansion of the datastore.
\item We elaborately design the specific key and value in the datastore with the fused multimodal information and reranking mechanism for more robust retrieval, which are novel and effective.
\item The experiment results show that our method\footnote{\url{https://github.com/Arvid-pku/RERIC}} achieves superior performance on the SIGHAN datasets compared with previous work.  
\end{enumerate}

\section{Backgrounds}
\subsection{Nearest Neighbor Language Model}
\label{2KNN}
Given a context sequence $\mathbf{c}_{t-1} = (w_1, \cdots, w_{t-1})$, the standard language model estimates the distribution over the next target token as $p_{\mathrm{LM}}\left(w_t\mid \mathbf{c}_{t-1}\right)$.
\citet{khandelwal2019generalization} proposed KNN-LM to involve augmenting the pretrained LM with the top-k nearest neighbors retrieval mechanism.

Firstly, let $f(\cdot)$ be the function that maps the context $\mathbf{c}_{i-1}$ to a fixed-length vector representation as the key, and the target token $w_i$ serves as the value.
Therefore, the training set $\mathcal{D}$ can be used to build the datastore for retrieval as:
\begin{equation}
(\mathcal{K}, \mathcal{V})=\left\{\left(f\left(\mathbf{c}_{i-1}\right), w_i\right) \mid\left(\mathbf{c}_{i-1}, w_i\right) \in \mathcal{D}\right\}
\end{equation}

Then at step $t$ during inference, given the context $\mathbf{c}_{t-1}$, the model queries the datastore with $f(\mathbf{c}_{t-1})$ to retrieve its $k$-nearest neighbors $\mathcal{N}$ using a distance function $d(\cdot, \cdot)$ and then obtains the additional probability of target token $w_t$ over the vocabulary:
\begin{align}
& p_{\mathrm{KNN}}(w_{t} \mid \mathbf{c}_{t-1})  \propto \notag \\
 & \quad \quad \sum_{\left(k_i, v_i\right) \in \mathcal{N}} \mathbb{I}_{w_t=v_i} \exp \left(-d\left(k_i, f(\mathbf{c}_{t-1})\right)\right)
\end{align}

Finally, the distribution obtained by retrieval will be interpolated to the standard LM distribution:
\begin{align}
& p(w_t \mid \mathbf{c}_{t-1})  =\lambda p_{\mathrm{KNN}}(w_t \mid \mathbf{c}_{t-1})\notag\\
& \quad \quad \quad \quad \quad +(1-\lambda) p_{\mathrm{LM}}(w_t \mid \mathbf{c}_{t-1})
\end{align}

\subsection{Chinese Spelling Check}
\label{2CSC}
The goal of the standard CSC model is to learn the conditional probability $\mathbf{p}_{\mathbf{CSC}}\left(\mathbf{y} \mid \mathbf{x}\right)$ for correcting a sentence $\mathbf{x} = \{ x_1, \cdots, x_n\}$ which may include spelling errors to the corresponding correct one $\mathbf{y}=\{y_1, \cdots, y_n\}$.
Correction is typically performed on each position according to the probability $p_{\mathrm{CSC}}\left(y_i \mid x_i, \mathbf{x}\right)$, which is based on the encoded representation of token $x_i$ in the whole context $\mathbf{x}$.

We represent the semantic representation of $x_i$ encoded by the CSC model as $s_i$, which is traditionally used. Due to the characteristics of CSC, spelling errors always arise from phonetic or morphologic similarity and confusion. Therefore, current CSC models also obtain representations of the phonetic and morphologic information of $x_i$, which are denoted as $p_i$ and $m_i$, respectively.

The final representation $r_i$ is obtained through fusing these representations above with an appropriate approach, such as a gating mechanism $g$:
\begin{equation}
    \begin{aligned}
    r_i = g(s_i, p_i, m_i)
    \end{aligned}
\end{equation}

\begin{figure*}
\centering 
\includegraphics[width=0.99\textwidth]{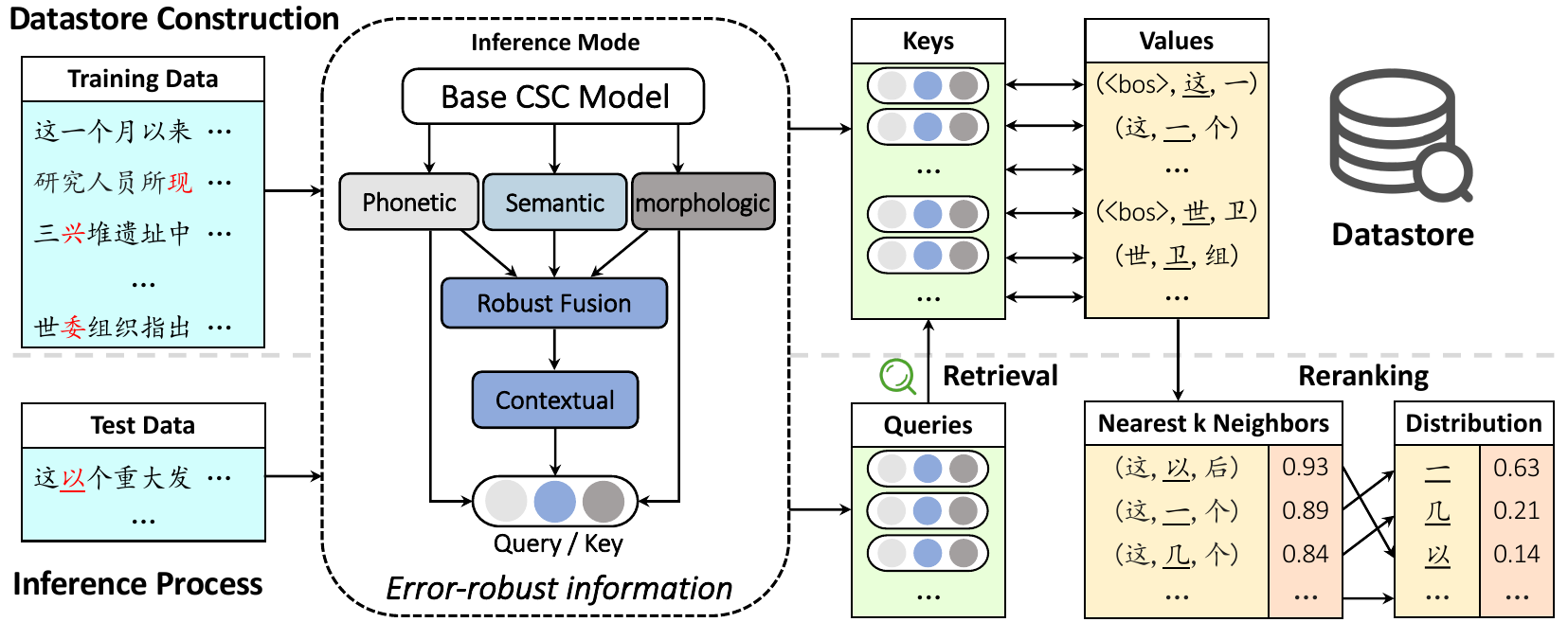} 
\caption{An illustration of our RERIC method with the datastore construction and the inference process including the KNN retrieval and reranking. The key contains the phonetic, morphologic, and contextual information of the token obtained from the base CSC model, and the value is in the form of 3-gram here. There are both correct (the majority) and incorrect tokens (marked in red) in the training and test data. Moreover, the target token and corresponding positions in n-gram values are underlined. And the test sample shows the correction process for the token "以(to)" , which should be corrected to "一(one)".
}
\label{Model}
\end{figure*}

Then, the CSC model will output probability distribution of each token over vocabulary $\mathcal{V}$ according to the corresponding fused representation $r_i$,
\begin{equation}
p_{\mathrm{CSC}}\left(y_i \mid x_i, \mathbf{x}\right)=t(r_i)
\end{equation}
where $t$ is a classification module, such as a linear transform followed by the softmax layer.


Since Chinese Spelling Check is usually performed by predicting each token with the corresponding representation, we can enhance the base CSC model with a k-nearest neighbor retrieval mechanism. Based on this idea, it is natural to construct a datastore using such representation vectors from the CSC model as dense indexes and then employ retrieval to integrate with the traditional CSC model. However, due to the characteristics and complications of the CSC task, such as the mix of correct and incorrect tokens in the input, we need to make necessary adjustments for error robustness, which will be described in Section \ref{method}.

\section{Methodology}
\label{method}
As shown in Figure \ref{Model}, the core idea of our work is to enhance the base CSC model with a novel k-nearest neighbor retrieval and n-gram reranking mechanism. The datastore built for retrieval is carefully designed according to the characteristics of the CSC task to improve robustness against potential spelling errors in the input text and also make better use of every token, no matter whether it is correct or not. In Section \ref{sec:build}, we introduce our design of key and value in the datastore construction. In Section \ref{sec:use}, we describe our reranking method and how to utilize retrieved KNN candidates.

\subsection{Datastore Construction}
\label{sec:build}
The structure of the datastore is a dictionary, in which each element consists of the pair \textit{(key, value)} based on the CSC training data. The key is used to retrieve the k-nearest neighbors for the target token, and the corresponding value serves as the probable candidates for further integration.

\paragraph{Key Design} 
The goal of our key design is to alleviate the negative impact of mixing correct and incorrect tokens in the input and provide sufficient information for error correction. Each target token needs to be represented more rationally and robustly in the same high-dimensional space.

We take the last sentence "世委组织指出...(The World Commission states that...)" in the training data in Figure \ref{Model} as an example. It should be corrected to "世卫组织指出...(WHO states that...)", where the token "委" is incorrect and has a similar pronunciation ("wei") to the correct token "卫". We believe that the pure semantic information of the incorrect token is misleading and unreliable for retrieval. In contrast, the phonetic or morphologic information is usually approximately correct in such cases, according to the study of \citet{liu2011visually}. Therefore, it is more reasonable to utilize multimodal information about pronunciation and morphology in the design of keys.

Furthermore, we extend the semantic information of the target token itself to the whole contextual information, which is included in the key calculation and is more robust and essential in CSC. In the example mentioned above, we indeed require the context of "委" that involves "世(world)" and "组织(organization)" to infer the correct expression "世卫组织(WHO)". Although the traditional semantic representation from language models has usually integrated contextual information to some extent, we suppose it is not robust enough because it is still mainly influenced by the target token. To further improve error-robustness, we propose a careful approach to obtaining the contextual representation.

Specifically, although the context contains more words so as to be possibly insensitive to individual potential incorrect tokens, such semantic errors still have negative impacts. So we fuse the phonetic and morphologic representations with the semantic ones when dealing with tokens in context. Moreover, we suppose that the nearer context content to the target token may play more important roles. Hence, we use a Gaussian distribution to calculate the weighted average of the fused representations of tokens around the current token to obtain our specific contextual representation.

Formally, we use unified definitions the same as in Section \ref{2CSC}, where $(\mathbf{x}, \mathbf{y})$ is denoted as a sample in the training data $\mathcal{D}$. And $s_t, p_t, m_t$ represent the semantic, phonetic, and morphologic representations of the target token $x_t$ at step $t$, respectively. In the calculation of contextual representation $c_t$, these representations of each token $x_i$ in the context are fused with the fusion function $g$, which is usually a gate mechanism. Then $c_t$ is obtained by weighting and averaging the representations of the neighboring tokens according to the Gaussian distribution $f_n$ and token distance $|i - t|$:
\begin{align}
    c_t &= \sum_{1\le i\le L}{f_n(i;t, \sigma)g(s_i, p_i, m_i)} \\
    f_n(i;t, \sigma) &= \frac{1}{\sqrt{2\pi}\sigma} exp(-\frac{(i-t)^2}{2\sigma^2})
\end{align}

where $L$ is the length of input sequence $\mathbf{x}$.

In the end, we use concatenation to combine and store three parts of information and obtain the key that is consistent with our motivation, representing the error-robust information (ERI).
\begin{equation}
    k_t = [p_t; m_t; c_t]
\end{equation}

\paragraph{Value Design}
As mentioned in Section \ref{2KNN}, the standard KNN-LM only uses the corrected target token $y_t$ as the value if directly applied to the CSC task, which is relatively simple and improvable.
Considering another sample "这以个(this to)..." in the test data in Figure \ref{Model}, we can find that the candidate "这以后(after this)..." ranks the highest. But the corresponding target token "以" is the same as that of the input and incorrect, which will lead to the wrong correction. In such cases, the influence of continuous contextual representations is insufficient since several candidates all contain similar contexts. And the same phonetic and morphologic representations as the input query indeed result in misleading advantages in ranking.

In view of this, we propose to extend the single target token to the n-gram around it as the value for further matching and reranking. They are performed through the discrete comparison of tokens in the n-gram value with the corresponding ones in the input. And we can reduce the role of the target token in matching, namely the central token of the n-gram, to deal with the cases mentioned above. For example, "这一个(this one)..." should be more likely to be correct than "这以后(after this)...“ for the input "这以个(this to)..." with more overlapping context except "以". More details will be described in Section \ref{sec:use}.
 
To be specific, we store the n-gram with a window of size $n$ centered on the corresponding target token $y_{t}$ in the output for explicitly matching the contextual information, as follows:
\begin{equation}
    v_t = [y_{t-\lfloor n/2 \rfloor}, \cdots, y_{t}, \cdots, y_{t+\lfloor n/2 \rfloor}]
\end{equation}
In these ways, the datastore for retrieval is finally built from the whole training dataset:
\begin{equation}
    (\mathcal{K}, \mathcal{V}) = \bigcup_{(\mathbf{x},\mathbf{y})\in \mathcal{D}} \{(k_t, v_t), \forall (x_t, y_t) \in (\mathbf{x}, \mathbf{y})\}
\end{equation}

\subsection{Retrieval and Reranking}
\label{sec:use}

\paragraph{Retrieval}
During the process of correction, for each token $x_t$ in the input sentence, our RERIC method aims to predict the corresponding $y_t$. Considering the intention of our retrieval augmentation, we obtain the specific query $q_t = [p_t; m_t; c_t]$ in the same way as the calculation of the key described in Section \ref{sec:build}.

Then the query is used to retrieve the k-nearest neighbors in the datastore we constructed before based on the measure of similarity. We utilize the common $l_2$ distance, which will continue to participate in subsequent weighting. Formally, the obtained k-nearest neighbors are denoted as $N_t = \{(k^i, v^i), i\in \{1, 2, \cdots k\}\}$ and their corresponding distance as $D_t=\{d(q_t, k^i), i\in \{1, 2, \cdots k\}\}$, where $d(q_t, k^i)$ means $l_2$ distance between $q_t$ and $k^i$.

\paragraph{Reranking}
In the preceding sections, we have introduced our novel n-gram value for further reranking to improve the utilization of retrieved candidates. Specifically, the n-gram of input target token $x_t$ for overlap matching is obtained similarly as $g_t = [x_{t-\lfloor n/2 \rfloor}, \cdots , x_{t}, \cdots, x_{t+\lfloor n/2 \rfloor}]$ with a window of size $n$ centered on $x_t$. We calculate the modified distance $d_t^j$ of the retrieved $j$-th neighbor for $x_t$ as:
\begin{equation}
    \begin{aligned}
        \alpha_t^j &= \frac{\sum_{1 \leq i \leq n }{\mathbb{I}(v^j(i), g_t(i))w_i}}{n}, \\
        d_t^j &= (1-\alpha_t^j) d(q_t, k^j),
    \end{aligned}
\end{equation}
where $w_i$ is the customized weight if the two tokens at position $i$ of $v^j$ and $g_t$ are the same. Particularly, the weight of the central position of the n-gram is diminished in order to address cases where the target tokens in $v^j$ and $g_t$ are the same and incorrect, like "这以后(after this)..." mentioned in Section \ref{sec:build}. And $\alpha_t^j$ represents how much the retrieved n-gram overlaps with the input, which can measure their similarity. The larger the overlap is, the smaller the modified distance $d_t^j$ becomes, and then the higher the corresponding candidate will rerank.

\paragraph{Utilization}
With the above designs, the probability distribution over the vocabulary of the target output token $y_t$ based on the retrieved neighbors and reranking is computed as:
\begin{align}
&p_{\mathrm{RERIC}}(y_t \mid x_t, \mathbf{x}) \propto \notag\\
& \quad \quad \quad\sum_{(k^i, v^i)\in N_t}{\mathbb{I}(y_t=v^i(\lfloor n/2 \rfloor))\mathrm{exp}(\frac{-d_t^i}{T})},
\end{align}
where $T$ is the softmax temperature and $v^i(\lfloor n/2 \rfloor)$ is the central word of n-gram $v^i$. The final probability when predicting $y_t$ is calculated as the interpolation of two distributions with a hyperparameter $\lambda$:
\begin{equation}
\begin{aligned}
p(y_t \mid x_t, \mathbf{x}) &=\lambda p_{\mathrm{RERIC}}(y_t \mid x_t, \mathbf{x}) \\
&+(1-\lambda) p_{\mathrm{CSC}}(y_t \mid x_t, \mathbf{x})
\end{aligned}
\end{equation}
where  $p_{\mathrm{CSC}}$ indicates the vanilla distribution from a base CSC model.

\section{Experiments}
In this section, we introduce the details of our experiments, including the datasets, metrics for evaluation, baselines, and the main results we obtained. Then, in the next section, we conduct further analyses and discussions to verify the effectiveness of our method.

\subsection{Datasets}

\paragraph{Training Data} We follow previous works on CSC \citep{zhang2020spelling,liu2021plome, xu2021read,li2022past} and use the same training data, including the training samples from SIGHAN13 \citep{wu2013chinese}, SIGHAN14 \citep{yu2014overview}, SIGHAN15 \citep{tseng2015introduction} and the pseudo training data, denoted as Wang271K \citep{wang2018hybrid}. In addition, we randomly select 10\% of the training data during training as our verification set to select the best hyperparameters.

\begin{table}[t]
\centering
\resizebox{0.48\textwidth}{!}{%
\begin{tabular}{c|ccc}
\toprule
Dataset & \#Sent & \#Error & \#Error-pair  \\ \midrule
SIGHAN Training & 6126 & 8470 & 3318 \\ 
Wang271K & 271329 & 381962 & 22409 \\ \midrule
SIGHAN13 Test & 1000 & 1217 & 748 \\ 
SIGHAN14 Test & 1062 & 769 & 461 \\ 
SIGHAN15 Test & 1100 & 703 & 460 \\ \bottomrule
\end{tabular}%
}
\caption{Statistics of the SIGHAN and Wang271K used in our experiments. We report the number of sentences in the datasets (\#Sent), the number of misspellings contained (\#Error) and the number of different kinds of errors (\#Error-pair).}
\label{statistic}
\end{table}

\paragraph{Test Data} To guarantee fairness, we use the same test data as previous work, which are from the SIGHAN13/14/15 test datasets. It is noted that the text of the original SIGHAN dataset is in Traditional Chinese, so we use OpenCC to pre-process these original datasets into Simplified Chinese, which has been commonly applied in previous work \citep{wang2019confusionset, cheng2020spellgcn, zhang2020spelling}. Detailed statistics of the training and test data we used in our experiments are shown in Table \ref{statistic}.

\subsection{Evaluation Methods}
We evaluate the performance of our RERIC method with the sentence-level metrics that are commonly used in the existing work of CSC.
The results are reported at both the detection level and the correction level.
At the detection level, a sentence is considered correct if all spelling errors in the sentence are successfully detected.
At the correction level, the spelling errors not only need to be detected but also need to be corrected.
We report accuracy, precision, recall, and F1 scores at both levels, the same as in previous studies.

\subsection{Baseline Models}
For better comparison of our method, we selected several advanced baseline methods with the same experimental settings as ours:
\textbf{FASpell} designed by \citet{hong2019faspell} is a model that consists of a denoising autoencoder and a decoder. 
\textbf{SpellGCN} \citep{cheng2020spellgcn} integrates the confusion set to the correction model through GCNs to improve CSC performance. 
\textbf{PLOME} \citep{liu2021plome} is a task-specific pretrained language model to correct spelling errors. 
\textbf{REALISE} \citep{xu2021read} is a multimodal CSC model which captures and mixes the semantic, phonetic and morphologic information. 
\textbf{ECOPO} \citep{li2022past} is an error-driven contrastive probability optimization framework and can be combined with other CSC models.
In addition, popular large language models like ChatGPT \footnote{\url{https://chat.openai.com/}} have recently made great progress and performed well on various tasks and domains. Therefore, we also conduct experiments with ChatGPT on CSC, which will be discussed in Section \ref{chatgpt}.


\begin{table}[]
\centering
\resizebox{0.46\textwidth}{!}{%
\begin{tabular}{cc|cc}
\toprule
Parameter & Value &  Parameter & Value              \\ \midrule
$k$       & 12    & $T$                   & 50                 \\
$n$-gram  & 3     & $w$                           & (1.68, 0.68, 1.68) \\
$\lambda$ & 0.4   & $\sigma$     & 1                  \\ \bottomrule
\end{tabular}%
}
\caption{The hyperparameters of our method.}
\label{tab:my para}
\end{table}%

\begin{table*}[t]
\centering
\renewcommand\tabcolsep{2.5pt}
\resizebox{\textwidth}{!}{
\begin{tabular}{c|c|cccc|cccc}
\toprule
\multirow{2}{*}{Test Set} & \multirow{2}{*}{Model} & \multicolumn{4}{c|}{Detection Level} & \multicolumn{4}{c}{Correction Level} \\
& & Accuracy & Precision & Recall & F1 Score & Accuracy & Precision & Recall & F1 Score \\
  

\midrule

\multirow{5}{*}{SIGHAN13} 
& FASpell \citep{hong2019faspell} & 63.1 & 76.2 & 63.2 & 69.1 & 60.5 & 73.1 & 60.5 & 66.2 \\
& SpellGCN \citep{cheng2020spellgcn} & - & 80.1 & 74.4 & 77.2 & - & 78.3 & 72.7 & 75.4 \\
& ECOPO$^{\dag}$ \citep{li2022past} & \textbf{83.3} & 89.3 & \textbf{83.2} & \textbf{86.2} & \textbf{82.1} & 88.5 & \textbf{82.0} & 85.1 \\
\cmidrule(lr){2-10} & REALISE$^{\dag}$ \citep{xu2021read} & 82.7 & 88.6 & 82.5 & 85.4 & 81.4 & 87.2 & 81.2 & 84.1 \\
& RERIC$^{\dag}$ (Ours) & 83.0 & \textbf{89.7} & 82.8 & 86.1 & \textbf{82.1} & \textbf{88.7} & 81.9 & \textbf{85.2} \\

\midrule

\multirow{5}{*}{SIGHAN14} 
& FASpell \citep{hong2019faspell} & 70.0 & 61.0 & 53.5 & 57.0 & 69.3 & 59.4 & 52.0 & 55.4 \\
& SpellGCN \citep{cheng2020spellgcn} & - & 65.1 & 69.5 & 67.2 & - & 63.1 & 67.2 & 65.3 \\
& ECOPO \citep{li2022past} & 79.0 & 68.8 & \textbf{72.1} & 70.4 & 78.5 & 67.5 & \textbf{71.0} & 69.2 \\
\cmidrule(lr){2-10} & REALISE \citep{xu2021read} & 78.4 & 67.8 & 71.5 & 69.6 & 77.7 & 66.3 & 70.0 & 68.1 \\
& RERIC (Ours) & \textbf{79.9} & \textbf{72.1} & 70.6 & \textbf{71.3} & \textbf{79.6} & \textbf{71.3} & 69.8 & \textbf{70.6} \\

\midrule

\multirow{6}{*}{SIGHAN15} 
& FASpell \citep{hong2019faspell} & 74.2 & 67.6 & 60.0 & 63.5 & 73.7 & 66.6 & 59.1 & 62.6 \\
& SpellGCN \citep{cheng2020spellgcn} & - & 74.8 & 80.7 & 77.7 & - & 72.1 & 77.7 & 75.9 \\
& PLOME \citep{liu2021plome} & - & 77.4 & 81.5 & 79.4 & - & 75.3 & 79.3 & 77.2 \\
& ECOPO \citep{li2022past} & 85.0 & 77.5 & \textbf{82.6} & 80.0 & 84.2 & 76.1 & \textbf{81.2} & 78.5 \\

\cmidrule(lr){2-10} & REALISE \citep{xu2021read} & 84.7 & 77.3 & 81.3 & 79.3 & 84.0 & 75.9 & 79.9 & 77.8 \\
& RERIC (Ours) & \textbf{86.1} & \textbf{81.1} & 81.3 & \textbf{81.2} & \textbf{85.6} & \textbf{79.9} & 80.1 & \textbf{80.0}\\

\bottomrule
\end{tabular}
}
\caption{Sentence-level performance of our RERIC method and baseline models. REALISE is the backbone and base CSC model for RERIC to build the datastore. Results marked with "\dag" on SIGHAN 2013 are post-processed with removing all "的", "地", "得" from the model output, due to the low annotation quality about them, which is to follow the previous work \citep{xu2021read} for convenient comparison.
}
\label{allres}
\vspace{-0.1in}
\end{table*}

\subsection{Implementation Details}

To get the error-robust information proposed to construct our datastore, we consider using a pretrained model on the CSC task, which is similar to the retrieval-related works in other domains.
And we choose REALISE, a strong multimodal model that captures semantic, phonetic, and morphologic information, which meets our requirements.

More specifically, a pretrained GRU encodes the pronunciation (pinyin) sequence of input to obtain the phonetic information $p_t$ of each token $x_t$, while a pretrained ResNet encodes the character graphics to obtain the morphologic information $m_t$.
Given the trade-off between storage space and model performance, we store 3-grams centered on the position of the current token as the value of the datastore.
We implement the grid search on the validation set to determine the hyperparameters of our experiments, and more details are shown in Table \ref{tab:my para}. Since we do not need to retrain the CSC model, we only focus on the hyperparameters during our retrieval-augmented inference.


\subsection{Experimental Results}
The main results on sentence-level metrics of our RERIC method and all baseline models are shown in Table \ref{allres}.
It can be observed that our method has obtained substantial improvements on SIGHAN14 and SIGHAN15 while achieving comparable results on SIGHAN13, compared to the previous state-of-the-art performance.
When turning to the REALISE, on which our method is based, the improvement is more remarkable, with an average increase of about 2.0\% on three SIGHAN test datasets.

On the other hand, it is notable that both the accuracy and precision scores of our method have improved remarkably, while the recall score has not shown much change.
It demonstrates that RERIC becomes less prone to wrong corrections, which may be due to the fact that the related correct samples in training data are better utilized.
More detailed analyses are provided in Section \ref{sec: analysis}.


\section{ Analyses and Discussions}
\label{sec: analysis}

\subsection{Ablation Experiments}
We conduct ablation experiments to analyze and verify the effects of different components of RERIC, involving error-robust information (ERI), n-gram value reranking (NVR), and specific contextual representation. The results are shown in Table \ref{tab:ablation}.

\begin{table}[t]
\centering
\renewcommand\tabcolsep{5pt}
\resizebox{0.48\textwidth}{!}{%
\begin{tabular}{lcccccc}
\toprule
\multirow{2}{*}{Method} & \multicolumn{3}{c}{Detection Level} & \multicolumn{3}{c}{Correction Level} \\ \cmidrule{2-7} 
           & Pre  & Rec  & F1  & Pre  & Rec  & F1  \\ \midrule
RERIC  & 81.1 & 81.3 & 81.2 & 79.9 & 80.1 & 80.0 \\
w/o ERI-P & 79.3 & 80.9 & 80.1 & 78.3 & 79.4 & 78.9 \\
w/o ERI-M & 79.5 & 81.0 & 80.2 & 78.7 & 79.5 & 79.1 \\
w/o ERI-C & 80.4 & 81.1 & 80.7 & 79.2 & 80.2 & 79.7 \\
w/ ERI-S & 74.6 & 78.4 &	76.5 & 72.0 & 75.7 & 73.8 \\
w/o ERI & 77.7 & 81.3 & 79.5 & 76.5 & 80.0 & 78.2 \\
w/o NVR & 79.7 & 81.2 & 80.4 & 78.4 & 79.9 & 79.1 \\
w/o Retrieval & 77.3 & 81.3 & 79.3 & 75.9 & 79.9 & 77.8 \\ \bottomrule
\end{tabular}
}
\caption{Ablation results of our RERIC method on SIGHAN2015 test set. We apply the following changes:
1) removing each component of ERI (w/o ERI-P, w/o ERI-M, and w/o ERI-C denote the reduction of phonetic, morphologic, and contextual information, respectively);
2) using traditional semantic representation as the component of ERI (w/ ERI-S);
3) using the standard hidden representation of the token as the key (w/o ERI);
4) removing the reranking process and only using the single token as the value (w/o NVR).}
\label{tab:ablation}
\vspace{-0.15in}
\end{table}

They indicate that all three types of representations in ERI are critical, especially the phonetic information.
It may be due to the fact that the lar gest proportion of errors in the SIGHAN test set are caused by phonetic similarity.
And there is a relatively small decrease when contextual information is removed, probably because the design of NVR also introduces contextual information into the model. We also show the results of the traditional usage of semantic representation, which perform much worse and prove the importance of our contextual representation.

Moreover, when ERI or NVR are removed, the performance of the model drops significantly at both detection and correction levels. The absence of reranking by the n-gram value may make the model unable to more robustly deal with the retrieved neighbors. Besides, using only the standard hidden representation of each token, like traditional retrieval methods, the model will be confused about whether the input token is reliable and almost fail to benefit from the retrieval augmentation. It proves that these components are crucial to our method and the improvement of the CSC task.

Moreover, in the case study, we find that the presence of similar samples in the training data and datastore successfully improves the model to avoid incorrect modifications. It is also consistent with the great increase in precision score of our RERIC method. Due to space limitations, we do not show detailed examples and illustrations here.


\subsection{Effects of Key Hyperparameters}
Despite the advantage of our method not requiring additional training, several key hyperparameters during inference and retrieval are crucial and should be determined carefully. Therefore, we show the effects of different hyperparameter settings in our method, together with the baseline scores from REALISE for convenient comparison.

\begin{figure}[t]
\centering 
\includegraphics[width=0.45\textwidth]{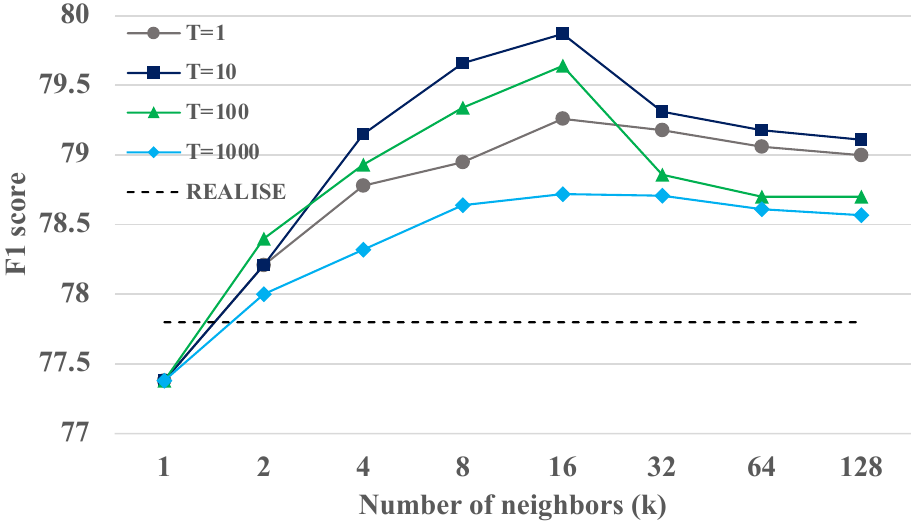} 
\caption{Effect of the number of retrieved neighbors $k$ and the softmax temperature $T$ on the SIGHAN 2015 test set. The performance of the baseline REALISE is represented as a dashed line.} 
\label{kt}
\end{figure}

\begin{figure}[t]
\centering 
\includegraphics[width=0.45\textwidth]{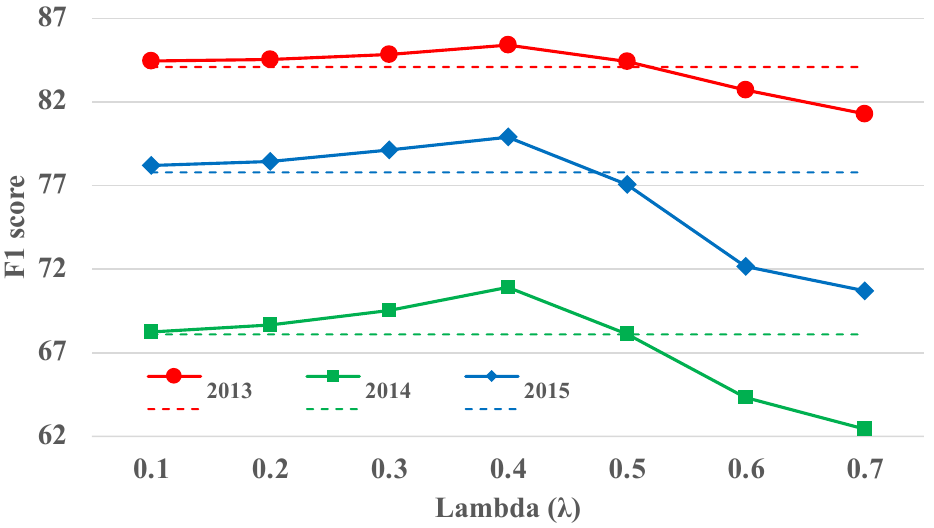} 
\caption{Effect of the interpolation parameter $\lambda$ on the SIGHAN 2013, 2014 and 2015 test set. The performance of the REALISE in three test sets is represented as the dashed line in the same color.} 
\label{ly}
\end{figure}

\noindent \textbf{Number of Neighbors $k$}$\quad$As shown in Figure \ref{kt}, initially, the performance improves with the increase in the number of neighbors used in the retrieval. But it starts to decrease when $k$ reaches about 16. It may be because more noise will be introduced if too many neighbors are utilized.

\noindent \textbf{Softmax Temperature $T$}$\quad$As shown in Figure \ref{kt}, the performance is relatively robust to softmax temperature $T$ and achieves good results over a wide range compared to the base model.

\noindent \textbf{Interpolation Parameter $\lambda$}$\quad$As shown in Figure \ref{ly}, our method performs best when $\lambda \approx 0.4$, which shows the optimal proportion for the interpolation of the base CSC model and predictions from retrieval.

\begin{table}
\centering
\resizebox{0.48\textwidth}{!}{
\begin{tabular}{ccccccccc}
\toprule
\multirow{2}{*}{Method} & \multicolumn{3}{c}{Detection Level} & \multicolumn{3}{c}{Correction Level} \\ \cmidrule{2-7} 
           & Pre  & Rec  & F1  & Pre  & Rec  & F1  \\ \midrule
ChatGPT & 36.8 & 79.4 & 50.3 & 26.5 & 57.2 & 36.2 \\ 
RERIC  & 81.1 & 81.3 & 81.2 & 79.9 & 80.1 & 80.0 \\
\bottomrule
\end{tabular}
}
\caption{Results of ChatGPT on the SIGHAN15 test set with the few-shot setting.}
\label{gptres}
\end{table}

\subsection{Discussions with ChatGPT}
\label{chatgpt}
To compare the currently popular and powerful large language models, we also conduct experiments on the CSC task and the SIGHAN15 test set with ChatGPT. We have attempted different input prompts for ChatGPT, such as instructions in English, multiple examples, explanations and reasons for error correction in the examples, etc. The best results we have obtained are shown in Table \ref{gptres}, using the setting of few-shot. However, we find that the results are, in fact, worse than many baselines, let alone our RERIC method. This is probably because ChatGPT is not adept at performing tasks like CSC that strictly restrict the output format, and thus there are many over-correction problems.

\begin{table}
\centering
\renewcommand\tabcolsep{2.5pt}
\resizebox{0.48\textwidth}{!}{
\begin{tabular}{ccccccccc}
\toprule
\multirow{2}{*}{Method} & \multicolumn{2}{c}{SIGHAN 2013} & \multicolumn{2}{c}{SIGHAN 2014} & \multicolumn{2}{c}{SIGHAN 2015} \\ \cmidrule{2-7} 
           & D-level & C-level & D-level & C-level & D-level  & C-level \\ \midrule
RERIC & 86.3 & 85.6 & 71.3 & 70.6 & 81.2 & 80.0 \\ 
+ DA & 86.5 & 85.6 & 71.7 & 70.7 & 81.5 & 80.9 \\
\bottomrule
\end{tabular}
}
\caption{The results of our RERIC method after performing data augmentation (DA). D-level and C-level denote the detection-level F1 score and correction-level F1 score, respectively.}
\label{dataaugtable}
\end{table}

\subsection{Data Augmentation}
In addition, we introduce specific data augmentation to expand our retrieval datastore, aiming to further improve the performance of our method. Notably, previous CSC studies often performed data augmentation by adding noise to raw texts with the confusion set obtained by rule-based approaches. However, such patterns cannot guarantee the quality of the synthetic data, which may consequently impair the performance. Moreover, they necessitate retraining the model, which consumes additional time and resources. In contrast, our method is more straightforward, allowing for the direct incorporation of raw and correct texts into the datastore without retraining or synthetic data construction. Specifically, we utilize the wiki2019zh dataset \citep{bright_xu_2019_3402023}, which comprises about one million articles from the Chinese Wikipedia without spelling errors. As demonstrated in Table \ref{dataaugtable}, after performing data augmentation, our method achieves an overall improvement, proving the efficacy of this simple strategy.

\begin{table}[]
\resizebox{0.48\textwidth}{!}{%
\begin{tabular}{ll}
\toprule
Input:          & 老师进教\textcolor[rgb]{1,0,0}{师}来了。          \\
Correct:        & 老师就进教\textcolor[rgb]{0,0.66,0.47}{室}来了。         \\
Translation: & \multicolumn{1}{l}{The teacher came into the classroom。}             \\
CSC Output:     & 老师就\textcolor[rgb]{1,0,0}{请}教\textcolor[rgb]{0,0.66,0.47}{室}来了。         \\
RERIC Output: & 老师就进教\textcolor[rgb]{0,0.66,0.47}{室}来了。         \\
Traing Sample:  & 当老师的第一个脚步踏进教室时... \\ \midrule
Input:          & 我\textcolor[rgb]{0,0.66,0.47}{带}上运动鞋出门。         \\
Correct:        & 我\textcolor[rgb]{0,0.66,0.47}{带}上运动鞋出门。         \\
Translation: & I take my sneakers and go out。                                       \\
CSC Output:     & 我\textcolor[rgb]{1,0,0}{戴}上运动鞋出门。         \\
RERIC Output: & 我\textcolor[rgb]{0,0.66,0.47}{带}上运动鞋出门。         \\
Traing Sample:                   & \begin{tabular}[l]{@{}l@{}}...带上半亿珠宝现身北京。\\ ...被老师带上街头。\end{tabular} \\ \bottomrule
\end{tabular}}
\caption{Some examples from SIGHAN 2015. The word in \textcolor[rgb]{1,0,0}{red} means an error, and the word in \textcolor[rgb]{0,0.66,0.47}{green} means correct. "CSC Output" means the prediction from standard REALISE model.} 
\label{case}
\end{table}

\subsection{Case Study}
It can be seen that the presence of similar contexts in the training set causes the model to prefer to keep the current token and therefore avoid incorrectly modifying it.
That's why in Table \ref{allres} the precision score of the model has increased a lot.

As shown in Table \ref{case}, given an input, "老师就进教师来了", which means "The teacher entered the teacher", the standard CSC model REALISE not only changes "师(teacher)" to "室(room)" but incorrectly changes "进(entered)" to "请(invite)".
Meanwhile the model argumented by the $k$NN avoids incorrect modifications successfully, benefit from a number of similar usages of "进" in the training set, such as ”当老师的第一个脚步踏进教室时" which means "When the teacher's first footsteps entered the classroom".

Another example is "我带上运动鞋出门(.I take my sneakers and go out.)", which is correct, but REALISE incorrectly changed the "带(take)" in it to "戴(wear)" which is usually used in Chinese to refer to putting on a hat, glasses, etc.
And RERIC does not make this mistake, because there are many similar uses of "带(take)" in the training set.

More similar examples can be found by comparing outputs of REALISE with our model.

\subsection{Inference Time}
We also investigate the influence of our RERIC method on the inference time of CSC through a comparison with the base CSC model REALISE. The results show that our retrieval method only causes a slight increase in inference time of only about 6\% with our hyperparameters and batch size of 32. Moreover, the storage increase is also small given the 1.3G size of the datastore. These issues are common problems with retrieval-related methods and have little impact on the actual implementation of the CSC task.

\section{Related Works}
\paragraph{Chinese Spelling Check}
CSC has received wide attention over the past decades.
Early work \citep{mangu1997automatic,jiang2012rule} used manually designed rules to correct the errors.
After that, methods based on statistical language models also made some progress \citep{yu2014chinese}.
With the development of deep learning and pretrained language model has achieved great improvements in recent years.
FASpell \citep{hong2019faspell} applied BERT as a denoising autoencoder for CSC. Soft-Masked BERT \citep{zhang2020spelling} chose to combine a Bi-GRU based detection network and a BERT based correction network.

In recent times, many studies have attempted to introduce phonetic and morphologic information into CSC models. SpellGCN was proposed to employ graph convolutional network on pronunciation and shape similarity graphs.
\citet{nguyen2020domain} employed TreeLSTM to get hierarchical character embeddings as morphologic information. 
REALISE \citep{xu2021read} used Transformer \citep{vaswani2017attention} and ResNet5 \citep{he2016deep} to capture phonetic and morphologic information separately.
In this respect，PLOME \citep{liu2021plome} chose to apply the GRU \citep{bahdanau2014neural} to encode pinyin and strokes sequence.
PHMOSpell \citep{huang2021phmospell} derived phonetic and morphologic information from multimodal pretrained models.

\paragraph{Retrieval Augmentation}
Retrieval-augmented text generation have been applied to many tasks including language modeling \citep{guu2020retrieval}, dialogue \citep{weston2018retrieve}, machine translation \citep{gu2018search} and others. 
\citet{li2022survey} provide an overview of this paradigm.

Of these retrieval-augmented methods, the studies that are most relevant to our work are KNN-LM \citep{khandelwal2019generalization}, which extends a pretrained language model by linearly interpolating it with a k-nearest neighbors model; KNN-NMT \citep{khandelwal2020nearest}, which combines the KNN algorithm closely with NMT models to improve performance; and BERT-KNN \citep{kassner2020bert}, which integrates the prediction from BERT for question with the KNN-based search.

\section{Conclusion}

In this paper, we propose RERIC to improve the current CSC model with our retrieval and reranking method. The key and value in the datastore for retrieval are elaborately designed according to the characteristics of CSC to effectively make use of the training data. More importantly, we employ multimodal representation that fuses phonetic, morphologic, and contextual information, together with n-gram matching and reranking, to improve error robustness during retrieval. The experimental results and relevant analyses prove the effectiveness of our method and its improvement over previous studies. Furthermore, our method can be simply applied in a plug-and-play manner without additional training, which shows superiority.




\bibliography{anthology,custom}
\bibliographystyle{acl_natbib}

\end{CJK*}
\end{document}